%% file: main.tex
\pdfoutput=1

\documentclass[11pt]{article}

\usepackage[final]{acl}

\usepackage{times}
\usepackage{latexsym}

\usepackage[T1]{fontenc}

\usepackage[utf8]{inputenc}

\usepackage{microtype}

\usepackage{inconsolata}

\usepackage{graphicx}
\usepackage{amsmath}
\usepackage{booktabs}
\usepackage{algorithm}
\usepackage{amssymb}
\usepackage{multirow}
\usepackage{algpseudocode}
\usepackage{caption}
\usepackage{subcaption}
\usepackage{makecell}
\usepackage{enumitem}
\usepackage{devanagari}
\usepackage{ulem}

\usepackage{titletoc}

\newcommand\corpusname{\textbf{\texttt{CoMuMDR}}}

\title{\corpusname: Code-mixed Multi-modal Multi-domain corpus for Discourse paRsing in conversations}

\author {
\textbf{Divyaksh Shukla}$^\clubsuit$  \qquad \textbf{Ritesh Baviskar}$^\clubsuit$ \qquad \textbf{Dwijesh Gohil}$^\diamond$ \\ \textbf{Aniket Tiwari}$^\diamond$ \qquad \textbf{Atul Shree}$^\diamond$ \qquad \textbf{Ashutosh Modi}$^\clubsuit$
\\
$^\clubsuit$Indian Institute of Technology Kanpur (IIT Kanpur)\qquad 
$^\diamond$Convin-AI\\
{\{divyaksh, ashutoshm\}}@cse.iitk.ac.in \qquad atul@convin.ai
}

\begin{document}
\maketitle
\input{sections/abstract}

\input{sections/introduction}
\input{sections/related-work}
\input{sections/datasets}
\input{sections/baseline-models}
\input{sections/discussion-conclusion}

\input{sections/limitations-ethics}

\bibliography{references}

\clearpage
\newpage

\input{sections/appendix}

\end{document}

%% file: sections/abstract.tex
\begin{abstract}

Discourse parsing is an important task useful for NLU applications such as summarization, machine comprehension, and emotion recognition. The current discourse parsing datasets based on conversations consists of written English dialogues    restricted to a single domain. In this resource paper, we introduce \textbf{CoMuMDR: Code-mixed Multi-modal Multi-domain corpus for Discourse paRsing} in conversations. The corpus (code-mixed in Hindi and English) has both audio and transcribed text and is annotated with nine discourse relations. We experiment with various SoTA baseline models; the poor performance of SoTA models highlights the challenges of multi-domain code-mixed corpus, pointing towards the need for developing better models for such realistic settings.  
\end{abstract}

%% file: sections/introduction.tex
\section{Introduction} \label{sec:introduction}

Discourse structures \cite{MANNTHOMPSON-rst-theory,Asher2005Jun-sdrt} capture relationships between clauses (e.g., causality, contrast, elaboration, and temporal sequencing) and are crucial to understanding the logical flow of information. These have been utilized in various tasks such as text summarization \cite{paulus2018a-summarization,li-etal-2016-role-summarization}, language understanding, machine reading comprehension \cite{li2019annotation-scheme-large-scale-multiparty}, dialog generation \cite{chernyavskiy-ilvovsky-2023-transformer,hassan-alikhani-2023-discgen,chen-yang-2023-controllable} and emotion recognition \cite{zhang-etal-2023-dualgats}. Researchers have created annotated discourse corpora from human-to-human dialogues for a single language such as English (e.g., STAC \cite{asher-etal-2016-discourse-stac} and Molweni \cite{li-etal-2020-molweni}). However, many modern-day conversations are audio-based and often involve code-mixing of multiple languages, such as Hindi and English (Hinglish). Understanding the discourse structure in such code-mixed audio-based conversations would be interesting. In this paper, we attempt to fill this gap. In a nutshell, we make the following contributions: 

\begin{figure}[t]
    \centering
    \includegraphics[scale=0.40]{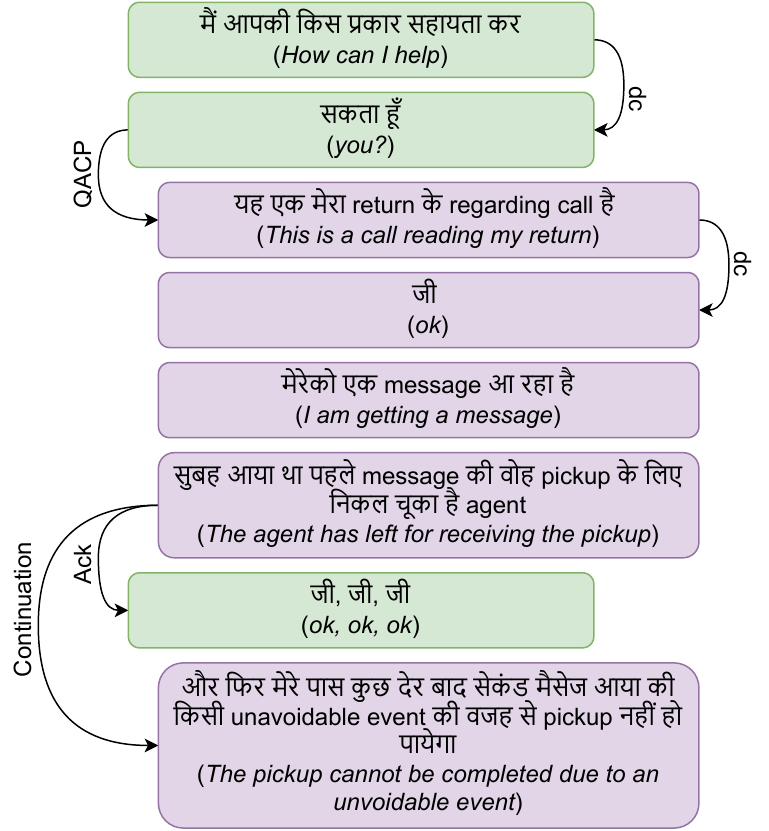}
    \caption{An example of a two-party call-center conversation (in code-mixed language) regarding a complaint on product return. Utterances corresponding to the customer center representative are shown in green boxes, and those of the customer are in purple boxes. Here, the relation types of \texttt{Question answer complaint pair} and \texttt{Acknowledgment} are shortened to \texttt{QACP} and \texttt{Ack}, respectively. Here, \texttt{dc} is the correction by the annotator on incorrect diarization.}
    \label{fig:thumbnail}
    \vspace{-6mm}
\end{figure}

\begin{itemize}[noitemsep,nosep,leftmargin=*]
\item We present \corpusname, a large scale code-mixed (Hindi + English = Hinglish), multi-modal (text+audio), multi-domain discourse corpus of two-party conversations (Table \ref{tab:dataset-comparison}). \corpusname\ consists of audio recordings and corresponding transcriptions of customer call center interactions from multiple domains, including e-commerce, pharmaceutical, stock broker application support, e-marketplace, and education. 
\item The corpus is annotated to create a labeled discourse graph for link prediction and discourse relation classification. The annotation is done at the span level with nine discourse relation types that aptly support the flow of information in customer call centers. We merged a few relation types presented in SDRT \cite{Asher2005Jun-sdrt} and added another type \texttt{Question answer complaint pair} to support the logical flow. Fig.  \ref{fig:thumbnail} shows a sample for \corpusname. The conversations in a practical setting can be complex; for example, there can be an overlap (\S \ref{sec:datasets}) between utterances (7th utterance) of two speakers. Also, note that since we used ASR and a diarization model for transcribing and splitting (\S \ref{sec:datasets}), an utterance (e.g., utterances 1, 2, and 3) could get incorrectly split due to diarization errors. These are resolved during annotations (\S \ref{sec:datasets}). 
\item We evaluate existing text-based discourse parsers (and GPT-4o) for link and relation prediction on \corpusname\ using English-only and multilingual text embeddings. We compare this with the performance of existing corpora, STAC and Molweni. We observe that SoTA models underperformed on \corpusname, pointing towards the need for the development of advanced models. 
\item We will release the experiment code, audio transcriptions (and text embeddings), and audio features via GitHub: \url{https://github.com/Exploration-Lab/CoMuMDR}. We do not release the actual audio and unfiltered transcripts due to concerns about the privacy of the company and its customers. 
\end{itemize}

\noindent The motivation behind \corpusname\ is to create a practical, real-world system that handles audio conversations and is robust to transcription and diarization errors.

%% file: sections/related-work.tex
\section{Related Work} \label{sec:related-work}

\begin{table}[t]
\tiny
\centering
\renewcommand{\arraystretch}{0.5}
\setlength\tabcolsep{3pt}
\begin{tabular}{@{}l lll@{}}
\toprule
                             & \textbf{STAC}    & \textbf{Molweni}    & \corpusname              \\
\midrule
\# dialogues                 & 1137             & 10000               & 799                      \\                   
\# utterances                & 10678            & 86042               & 8811                     \\
\# words                     & 44843            & 860851              & 79867                    \\
Avg. \#  utterances/dialogue & 11.07            & 8.83                & 11.03                    \\
Avg. \# words/dialogue       & 39.44            & 95.65               & 99.96                    \\
Parties                      & Multi            & Multi               & Two                      \\
Modalities                   & Uni-modal        & Uni-modal           & Multi-modal              \\
Languages                    & English          & English             & Code-mixed               \\
Source                       & Catan Game       & Ubuntu chats        & Call center interactions \\
Domains                      & Single domain    & Single domain       & Multi-domain             \\
Discourse Labels             & 17 labels        & 17 labels           & 9 labels                 \\
Annotator Metrics            & Kappa            & Kappa               & Kappa, Jaccard           \\
Data split \# dialogues      &                  &                     &                          \\
Train                        & 909              & 9000                & 639                      \\
Test                         & 115              & 500                 & 81                       \\
Validation                   & 113              & 500                 & 79                       \\
\bottomrule
\end{tabular}
\vspace{-3mm}
\caption{Comparison with previous corpora} 
\label{tab:dataset-comparison}
\vspace{-6mm}
\end{table}

Discourse Parsing has been an active research area in the NLP community \cite{li2022survey}. Discourse parsing consists of three main components: discourse segmentation \cite{wang-etal-2018-toward,lukasik-etal-2020-text,liu-etal-2021-dmrst}, discourse link prediction and discourse relation classification. Discourse segmentation divides a text corpus into Elementary Discourse Units (EDUs) for further processing. Discourse link prediction predicts a directed link between two EDUs, and relation classification assigns a relation type to the link (also check discourse theories in App. \ref{app:discourse-theory}). 

\noindent\textbf{Datasets:} In the context of English, two main text-based corpora have been proposed for Discourse parsing: \textbf{STAC} \cite{asher-etal-2016-discourse-stac}  and \textbf{Molweni} \cite{li-etal-2020-molweni} (check details in App. \ref{app:datasets}). Table \ref{tab:dataset-comparison} compares the STAC and Molweni datasets with our proposed dataset. \corpusname\ is code-mixed, audio-based, and covers multiple domains as opposed to mono-lingual single-domain conversations covered by existing text-based datasets. The corpus (having a comparable number of words with STAC) is based on Hindi-English code-mixed audio conversations with imperfect transcription and diarization quality, so \corpusname\ proposes a practical outlook on discourse parsing in conversations. Note that compared to existing datasets (STAC (based on the Catan game) and Molweni (based on Ubuntu chats)), \corpusname, besides including audio information, covers more domains and a variety of topics. 

\noindent\textbf{Discourse Parsing Models:} Various approaches have been proposed for Discourse parsing such as deep sequential model \cite{deep-sequential}, hierarchical model \cite{liu-chen-2021-improving}, Structure-aware model \cite{ijcai2021p543-structure-self-aware}, SSP-BERT+SCIJE model by \citet{yu-etal-2022-speaker} and SDDP model by \cite{chi-rudnicky-2022-structured}. Due to space constraints, details are given in App. \ref{app:related-model}. We benchmark using each of the above models.

%% file: sections/datasets.tex
\section{\corpusname\ Creation} \label{sec:datasets}

\begin{figure}[t]
 \centering
 \includegraphics[width=\columnwidth]{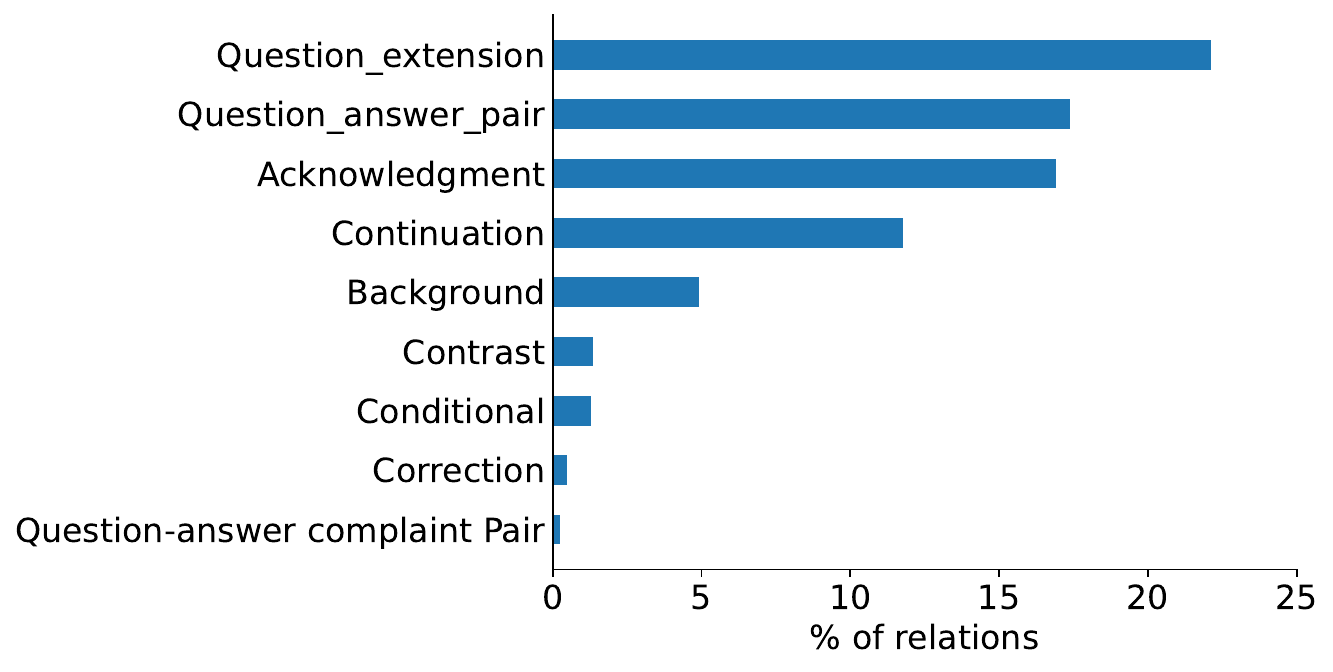}
 \vspace{-6mm}
 \caption{Distribution of discourse labels in \corpusname.}
 \label{fig:ours-label-distribution}
 \vspace{-6mm}
\end{figure}
 
\corpusname\ consists of two-party customer call center interactions. We obtained the data via a joint research collaboration with a call center company (they own the data) and want to automate customer call understanding. The calls mainly cater to Indian customers and companies. We ensure that the privacy of customers and companies mentioned in a call is maintained during annotation. The audio data is transcribed using the existing ASR (Automatic Speech Recognition) system \citep{verma23-interspeech-asr} (details in App. \ref{app:corpus-asr}) and diarized into utterances \citep{koluguri-2021-titanet-neural-model-speaker} (details in App. \ref{app:corpus-diarization}). Subsequently, the data is anonymized for customer names and other private information. A team of three  professional annotators further annotated the transcribed and diarized data. 

\noindent\textbf{Discourse Labels:} 
We used nine discourse labels: \texttt{Acknowledgment, Question-Answer pair, Question-Answer Complaint pair, Background, Contrast, Correction, Question Extension, Conditional,} and \texttt{Continuation}.  App. Table \ref{tab:discourse-labels} lists the discourse labels (and definitions). The labels are based on Semantic Discourse Representation Theory (SDRT) \cite{asher-etal-2016-discourse-stac}. App. \ref{app:distance-edus} provides details of the distribution of relative distance between linked EDU pairs for each relationship type (discourse label). In addition, we propose a new relation type, \texttt{Question-Answer Complaint Pair,} to classify complaints separately. We removed the \texttt{Narration} discourse label as it was not required in two-party customer conversations. During a pilot annotation experiment, we found that several discourse labels conveyed overlapping meanings. Hence, we merged them to get nine discourse labels to annotate our data (see Fig. \ref{fig:ours-label-distribution}). We observe a shift in the distribution of the labels when compared with STAC and Molweni. Specifically, \corpusname\ has a lower frequency of \texttt{Continuation} relation and a higher frequency of \texttt{Question Answer pair} and \texttt{Acknowledgment} relations, which is typical of customer call center interactions (see App. Fig. \ref{fig:stac-molweni-label-distribution}).

\noindent \texttt{Question Extension} label is arrived by merging \texttt{Clarification Question} and \texttt{Question Elaboration}, as all the instances of customer call center conversations posed clarification questions as elaborations, and the answers to them were more akin to answers to elaborative questions. \texttt{Conditional} is made by merging \texttt{Alternation} and \texttt{Conditional} as due to the nature of Hindi-English code-mixed conversations, it is hard to differentiate between a conditional and an alternation. 
\texttt{Continuation} is made by merging \texttt{Comment}, \texttt{Elaboration}, \texttt{Parallel} and \texttt{Result} since call center conversations rarely contain examples of comments compared to STAC and Molweni. Customer calls do not revolve around multiple ideas or topics; hence, there is no notion of parallel. Customer call center representatives continuously assure the customers about quick resolution of complaints, and the result of the conversation is typically implicit. 

\noindent Dialogues are divided into utterances using a diarization model. However, the audio contains overlapping utterances where both speakers are speaking simultaneously. Hence, we added another relation termed \texttt{Diarization Continuation} to fix the diarization issues. However, this relation label is not part of the discourse and is not used in calculating the results (\S\ref{sec:baselines}). We plan to use these annotations to improve the diarization model. 

\noindent\textbf{Annotation Details:} Annotators were tasked to identify the Elementary Discourse Units (EDUs), predict links between them to form a DAG, and assign a label to the relation between EDUs. A team of two annotators independently annotated the data, and another annotator verified the annotations and marked batches for re-annotation if deemed necessary. Previous studies have shown that identifying Complex Discourse Units (CDUs)\footnote{Multiple EDUs are combined using discourse relations to form a Complex Discourse Unit (CDU), making a discourse tree structure.} is a challenging task and have relied on combining EDUs using discourse relations to get a CDU \cite{muller-etal-2012-constrained,afantenos-etal-2015-discourse}. Prior discourse parsing models have used various strategies to convert CDUs to EDUs for efficient parsing \cite{deep-sequential, liu-chen-2021-improving}.  
We identified similar challenges in annotation 
and hence, instructed the annotators to connect an EDU with only the head (i.e., the first EDU) of a CDU \cite{asher-etal-2016-discourse-stac}. Appendix \ref{sec:annotation-details} presents more details regarding the annotation. We used the Inception software \cite{klie-etal-2018-inception} for annotation (\S \ref{app:anno-software}). We provide details of annotators,  instructions, and processes in the App. \ref{app:anno-profile}, App. \ref{app:anno-insturctions}, and App. \ref{app:anno-process}, respectively.

\noindent\textbf{Inter-Annotator Agreement:} Table \ref{tab:annotation-metrics} shows inter-annotator agreement using various metrics. Given our complex setting, existing metrics (e.g., Kappa) show a relatively low performance compared to previous datasets. We computed Kappa \cite{kappa-McHugh2012Oct} using the span and relation exact match metrics as in STAC and Molweni \cite{asher-etal-2016-discourse-stac,li-etal-2020-molweni}.


\begin{table}[t]
    \centering
    \tiny
    \begin{tabular}{lc}
    \toprule
        \textbf{Metric} & \textbf{Score}\\
        \midrule
        Jaccard & 0.9569\\
        Span exact match (see App. \ref{app:anno-metric})& 0.6294 \\
        Span partial match (see App. \ref{app:anno-metric})& 0.8224 \\
        Relation exact match (see App. \ref{app:anno-metric})& 0.5500 \\
        Relation partial match (see App. \ref{app:anno-metric}) & 0.5321 \\
        Structured Kappa \cite{asher-etal-2016-discourse-stac,li-etal-2020-molweni} & 0.4044 \\
        Relationship Kappa \cite{asher-etal-2016-discourse-stac,li-etal-2020-molweni} & 0.3190 \\
        \bottomrule
    \end{tabular}
    \vspace{-3mm}
    \caption{Inter-annotator agreement metrics} 
    \label{tab:annotation-metrics}
    \vspace{-6mm}
\end{table}

%% file: sections/baseline-models.tex
\section{Experiments, Results and Analysis} 
\label{sec:baselines}

\noindent\textbf{Discourse Modeling:} A dialogue consists of a list of utterances between two speakers. The utterances are further divided into elementary discourse units (i.e., clauses \cite{Asher2005Jun-sdrt}) $\{u_0, u_1, ..., u_n\}$, where $u_0$ is a dummy root EDU. Discourse parsing involves predicting a directed link between two EDUs $u_j$ and $u_i$ and assigning a relation label $r_{ji}$ between EDUs $u_j$ and $u_i$.

\noindent\textbf{Experimental Setup:} We experimented with state-of-the-art discourse parsing models: deep sequential model \cite{deep-sequential}, hierarchical model \cite{liu-chen-2021-improving}, Structure-aware model \cite{ijcai2021p543-structure-self-aware}, SSP-BERT+SCIJE model \citep{yu-etal-2022-speaker} and SDDP model \citep{chi-rudnicky-2022-structured}. We implemented all the discourse parsing models on STAC, Molweni, and \corpusname\  and trained them from scratch (details in App. \ref{app:hyper-params}). 
We followed the data-split (train/validation/test) as given in Table \ref{tab:dataset-comparison}. Validation set was used to tune the models. We implemented the models in two settings depending how texts in EDUs are encoded: English-only and multilingual embeddings. English-only embeddings include GLoVe \cite{pennington-etal-2014-glove} or \texttt{Roberta-base} embeddings \cite{liu-etal-2019-roberta}, same as those used in the original implementations. On the other hand, multilingual sentence-level embeddings include \texttt{paraphrase-xlm-r-multilingual-v1} \cite{reimers-gurevych-2019-sentence-bert}, which convert a complete EDU's text to a 768-dimension vector. 



\begin{table}[t]
\centering
\resizebox{\columnwidth}{!}{
    \begin{tabular}{@{}l ccc ccc@{}}
    \toprule
         & \multicolumn{3}{c}{\textbf{Link only}}& \multicolumn{3}{c}{\textbf{Link + relation}} \\ 
         & STAC & Molweni & \corpusname\ & STAC & Molweni & \corpusname\ \\ 
         \midrule
        \multicolumn{7}{c}{Multi-lingual embeddings} \\
        Hierarchical        & 0.6841 & 0.7000 & 0.9036 & 0.5221 & 0.5733 & 0.4263 \\ 
        Structure-aware     & 0.7125 & 0.8050 & \underline{0.9530} & 0.5314 & 0.5614 & 0.4850 \\
        SSP-BERT + SCIJE    & 0.7250 & 0.8205 & \textbf{0.9531} & \underline{0.6151} & \textbf{0.6634} & 0.5547 \\ 
        SDDP                & 0.7304 & 0.7898 & 0.9416 & 0.5670 & 0.5770 & 0.3781 \\
        \midrule
        \multicolumn{7}{c}{English-only embeddings} \\
        Deep Sequential     & 0.7496 & 0.7577 & 0.7330 & \textbf{0.6318} & 0.5162 & 0.4796 \\ 
        Hierarchical        & 0.7505 & 0.8097 & \underline{0.9443} & 0.5704 & 0.5690 & 0.5786 \\ 
        Structure-aware     & 0.7267 & 0.8232 & 0.7782 & 0.5582 & \underline{0.5934} & 0.4072 \\
        SSP-BERT + SCIJE    & 0.7201 & 0.8293 & \textbf{0.9452} & 0.5623 & 0.5925 & 0.5675 \\ 
        SDDP                & 0.7488 & 0.8233 & 0.7918 & 0.5887 & 0.5770 & 0.2941 \\
        \bottomrule
    \end{tabular}
    }
    \vspace{-3mm}
    \caption{F1-score of various discourse parsing models.  Values in \textbf{bold} highlight the top-performing model in each method, while values in \underline{underline} highlight the next top-performing model.}
    \label{tab:baseline-performance-xlm}
    \vspace{-2mm}
\end{table}

\begin{figure}
    \centering
    \includegraphics[scale=0.50]{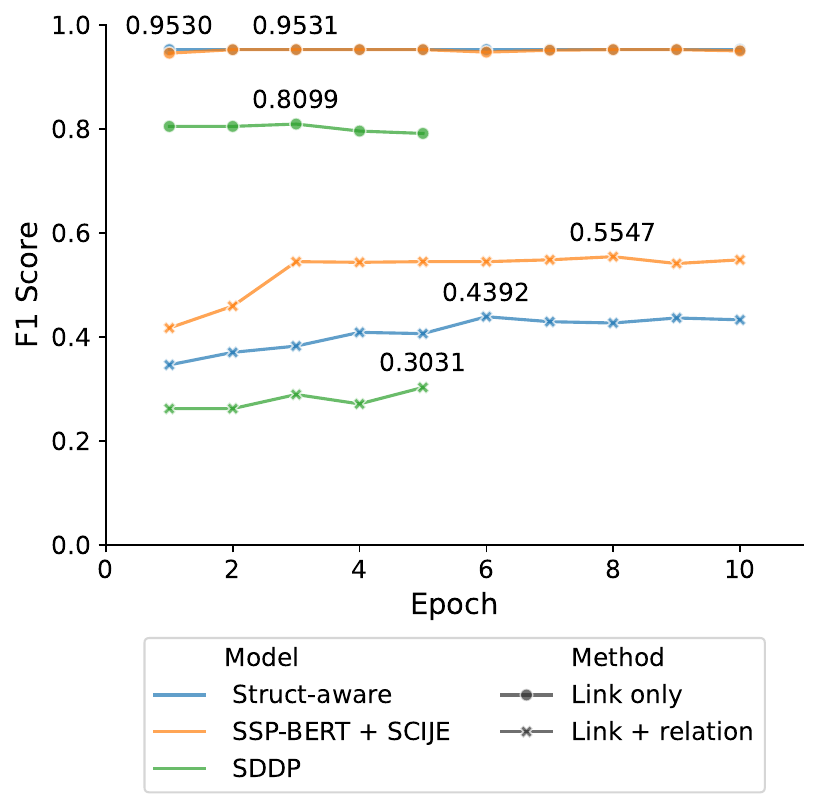}
    \vspace{-3mm}
    \caption{Comparing SDDP, Struct-Aware, and SA-DPMD across epochs on \corpusname\ test set. The values on each plot indicate the highest F1 score with respect to each model and discourse parsing method.}
    \label{fig:epoch-results}
    \vspace{-5mm}
\end{figure}

\begin{table*}[t]
\resizebox{\textwidth}{!}{
    \centering
    \begin{tabular}{lccc ccc ccc ccc c}
        \toprule
        & \multicolumn{3}{c}{\textbf{Hierarchical}} & \multicolumn{3}{c}{\textbf{Struct-Aware}} & \multicolumn{3}{c}{\textbf{SSP-BERT + SCIJE}} & \multicolumn{3}{c}{\textbf{SDDP}} & \\
        Label & Precision & Recall & F1-score & Precision & Recall & F1-score & Precision & Recall & F1-score & Precision & Recall & F1-score & Support \\
        \midrule
        \texttt{Question Extension}        & 0.39 & 0.47 & 0.42 & 0.00 & 0.00 & 0.00 & 0.57 & 0.61 & 0.59 & 0.18 & 0.30 & 0.22 & 172 \\
        \texttt{Acknowledgment}            & 0.62 & 0.68 & \textbf{0.65} & 0.56 & 0.47 & \textbf{0.51} & 0.68 & 0.72 & \textbf{0.70} & 0.04 & 0.50 & 0.08 & 151 \\
        \texttt{Question Answer}           & 0.54 & 0.58 & 0.56 & 0.40 & 0.66 & 0.49 & 0.53 & 0.45 & 0.49 & 0.00 & 0.00 & 0.00 & 146 \\
        \texttt{Continuation}              & 0.38 & 0.43 & 0.40 & 0.31 & 0.44 & 0.36 & 0.38 & 0.47 & 0.42 & 0.00 & 0.00 & 0.00 & 109 \\
        \texttt{Background}                & 0.00 & 0.00 & 0.00 & 0.00 & 0.00 & 0.00 & 0.28 & 0.22 & 0.24 & 1.00 & 0.11 & 0.20 & 32 \\
        \texttt{Contrast}                  & 0.00 & 0.00 & 0.00 & 0.17 & 0.08 & 0.11 & 0.50 & 0.44 & 0.47 & 0.40 & 0.39 & \textbf{0.39} & 8 \\
        \texttt{Conditional}               & 0.00 & 0.00 & 0.00 & 0.14 & 0.07 & 0.09 & 0.00 & 0.00 & 0.00 & 0.00 & 0.00 & 0.00 & 6 \\
        \texttt{Question Answer Complaint} & 0.00 & 0.00 & 0.00 & 0.00 & 0.00 & 0.00 & 0.00 & 0.00 & 0.00 & 0.00 & 0.00 & 0.00 & 3 \\
        \texttt{Correction}                & 0.00 & 0.00 & 0.00 & 0.00 & 0.00 & 0.00 & 0.00 & 0.00 & 0.00 & 0.00 & 0.00 & 0.00 & 2 \\
        \bottomrule
    \end{tabular}
    }
    \vspace{-3mm}
    \caption[Relation prediction error analysis]{Comparing precision, recall and F1 scores of predicted relations across three baseline discourse parsing models with multi-lingual embeddings on the test set of \corpusname.}
    \label{tab:relation-scores}
    \vspace{-5mm}
\end{table*}


\noindent\textbf{Results:} Table \ref{tab:baseline-performance-xlm} shows the F1-score (App. \ref{app:eval-metric}) on test sets for link and link+relation prediction. 
For both the settings, \corpusname\ scores are lowest across all the models, highlighting the challenge of discourse parsing on multi-domain, multilingual conversations. 
As can be observed, \corpusname\ has an equivalent score across models for link prediction. SDDP and Hierarchical model outperform on \corpusname\ compared to STAC and Molweni. However, in relation classification (link+relation), \corpusname\ has the lowest performance, possibly due to the presence of multiple domains and the challenge of domain adaptation \cite{liu-chen-2021-improving}. 

\noindent\textbf{Error Analysis:} 
In Table \ref{tab:baseline-performance-xlm}, we have computed the results of link and link+relation prediction of SDDP using Roberta and \texttt{paraphrase-xlm-r-multilingual-v1}. Roberta handles English-only text and cannot handle code-mixed or Hindi text. On the other hand, \texttt{paraphrase-xlm-r-multilingual-v1} can handle multilingual text but often fails at effectively processing code-mixed text. Hence, there is lower performance on relation prediction for SDDP. Baseline methods, including Deep sequential, hierarchical, Structure-aware, SSP-Bert + SCIJE, perform relation prediction after link prediction, i.e., they classify the relation type for each predicted link. On the other hand, SDDP performs link+relation prediction simultaneously as a single task, which is much more complicated. Hence, SDDP shows significantly lower performance on link+relation prediction than other baseline methods. Additionally, SDDP assumes the discourse relations to form a tree and performs tree parsing during inference, while most of the discourse relations in \corpusname\ cannot adhere to tree structures. Hence, SDDP on \corpusname\ shows low scores on link+relation prediction.

\noindent We further analyzed the learning dynamics of the baseline models on \corpusname\ and present results in Fig. \ref{fig:epoch-results}. The utterances were encoded using \texttt{paraphrase-xlm-r-multilingual-v1} while training all models on \corpusname. For the structure-aware and SSP-BERT + SCIJE models, we observed a rapid increase in F1-scores for link prediction, suggesting that the learned representations generalize well just after a single epoch. Notably, the F1-score for the struct-aware model remains nearly constant over 10 epochs, indicating early saturation in learning for link prediction. In contrast, link+relation prediction requires multiple epochs to generalize, and the corresponding scores exhibit high variance across epochs. This suggests that while the model may be well-suited for link prediction, it struggles to maintain stability and consistency for the more complex link+relation task.

\noindent Despite achieving notable improvements in link prediction on \corpusname\ using multilingual embeddings, we observe only a marginal increase in link+relation prediction scores. Table \ref{tab:relation-scores} provides a breakdown of precision, recall, and F1-scores for relation type prediction across SSP-BERT + SCIJE, structure-aware, and SDDP models. These scores reflect the underlying relation distribution shown in Fig. \ref{fig:ours-label-distribution}. Relations with low support counts are not effectively learned—even when using a weighted loss function, where the weights are computed as the inverse of their support counts. This is evident from the zero F1-scores for several rare relation types across all three models. For example, struct-aware achieves an F1-score of zero on the \texttt{Question Extension} relation, which contributes to its reduced overall performance on CoMuMDR despite strong results on link prediction. Overall, SSP-BERT + SCIJE outperforms the other baselines on \corpusname, but still falls short in predicting rare relations such as \texttt{Conditional}, \texttt{Question answer complaint}, and \texttt{Correction}. On further analysis, we observed that the hierarchical model \cite{liu-chen-2021-improving} could not predict the same relation links for \texttt{Correction} and \texttt{Contrast} on \corpusname, leading to a loss of performance in link prediction and relation classification involving correction and contrast (see Table \ref{tab:relation-scores}). The hierarchical model easily identifies \texttt{Acknowledgment} relations among the correctly predicted links. It could be due to the strong presence ($\sim 18\%$) of \texttt{Acknowledgment} in the dataset. Similarly, in SSP-BERT, the model misclassified some \texttt{Acknowledgment} relations as \texttt{Question answer pairs}. App. Fig. \ref{fig:error-sample-1} is an example of a conversation snippet with the gold and predicted relations marked on the left and right sides, respectively. The model incorrectly classified an \texttt{Acknowledgment} relation as a \texttt{Question Answer pair}, possibly due to the presence of ``ma'am'' in the acknowledgment clause (also see App. Fig. \ref{fig:error-sample-1}).

\begin{table}[t]
    \centering
    \resizebox{0.6\columnwidth}{!}{
    \begin{tabular}{l cc}
    \toprule
        ~ & \textbf{Link only} & \textbf{Link+relation} \\ 
        \midrule
        STAC & 0.6012 & 0.2729 \\
        Molweni & 0.5176 & 0.1474 \\
        \corpusname & \textbf{0.7217} & \textbf{0.2808} \\ 
    \bottomrule
    \end{tabular}
    }
     \vspace{-1mm}
    \caption{Performance of GPT-4o as a discourse parser.} 
    \label{tab:llm-scores}
    \vspace{-5mm}
\end{table}

\noindent\textbf{Results of GPT-4 Model:} We evaluated GPT-4o on the test set (81 dialogues, 890 utterances). We prompted GPT-4o in a 3-shot setting (template in App. \ref{app:gpt-4-template}) to behave as a discourse parser (results in Table \ref{tab:llm-scores}). GPT-4o performs worse on both tasks compared to the SoTA models. Upon examining the confusion matrix (App. Table \ref{tab:gpt-confusion-matrix}) for GPT-4o on \corpusname, we observed the misclassifications of \texttt{Question extension} as \texttt{Continuation}, possibly due to the overlapping semantics of these relations in a two-party conversation.

%% file: sections/discussion-conclusion.tex
\section{Future Directions and Conclusion} \label{sec:conclusion}

This paper presents \corpusname, a new discourse corpus for multi-modal, multi-domain, and code-mixed conversations from various customer call centers. We transcribed the audio and diarized the text into utterances. We annotated the EDUs using nine discourse labels by combining a few closely related labels from the SDRT format as they formed a more appropriate flow of discourse in a two-party conversation on customer support calls. In this work, we experimented with SoTA models; however, these do not perform well on \corpusname. In the future, we plan to develop more advanced models incorporating audio modality information. 


%% file: sections/limitations-ethics.tex
\section*{Limitations}

We developed \corpusname\ by capturing audio conversations between a customer and a customer care representative. The audio is then transcribed for annotation. 

\noindent Our corpus is not as big as the existing Discourse corpora but our corpus is code-mixed, multi-domain, and multi-modal. The corpus is sizable enough to develop meaningful models. Nevertheless, we plan to keep growing our corpus. Discourse annotations is a very time consuming process and hence it takes time to expand the corpus.   

\noindent \corpusname\ consists of nine discourse relation labels, far fewer than STAC and Molweni, which contain 17 labels. We found during our pilot annotation process that the \texttt{Narration} discourse label had no role in customer-centered conversations. Also, we found that in two-party conversations, some of the discourse labels had quite confusing meanings, which led to poor inter-annotator agreements. Hence, we combined the labels to create our presented nine labels presented in Table \ref{tab:discourse-labels}.

\noindent To build the dataset, we collected audio recordings from customer care centers. The audio was then transcribed and diarized. We found that the state-of-the-art diarization model gave imperfect diarizations during our pilot annotation process. It is because the audio data we collected consists of overlapping audio, i.e., both speakers are speaking simultaneously, and the transcription model returns text for both speakers. Hence, we added another annotation termed \texttt{diarization continuation}, and the annotators were tasked to fix the diarization issues along with discourse relation annotation. 

\noindent The RST and SDRT theories \cite{MANNTHOMPSON-rst-theory,Asher2005Jun-sdrt} define clauses as the textual span to be used as elementary discourse units (EDU). However, due to the nature of \corpusname\ and the imperfect diarizations resulting from the same, we could not use off-the-shelf clause identification algorithms. Hence, our annotation effort also includes the manual identification of EDUs and discourse relation annotation. It led to annotator-level differences in selecting clause spans. Hence, we report different annotation metrics in Appendix \ref{sec:annotation-details}.

\section*{Ethical Considerations} \label{sec:ethical-considerations}

\corpusname\ is constructed by obtaining audio conversation data from customer call center offices. The data is obtained under the agreement between us and the research collaborator (call center company). All the data that was used for experimentation complies with the terms of use and licensing agreements.

\noindent The audio transcriptions in \corpusname\ are anonymized for of all personally identifiable information. We also removed instances of toxic language, offensive or harmful content, and sensitive or wrong information from \corpusname.

\noindent \corpusname\ consists of Hindi-English code-mixed conversations taken from a specific geographical section. The data contains conversations from companies in pharmaceutical, e-commerce, stock broker applications, e-marketplaces, and education counseling services. 

\noindent We made sure to remove any bias in the data. Any bias, toxic language, offensive or harmful content, sensitive information, and misinformation in \corpusname\ is entirely unintentional.

\noindent Due to licensing agreements and ethical constraints, we will not be releasing the original audio data in \corpusname. We will only release the anonymized text transcriptions, corresponding text embeddings and audio features along with appropriate annotations in \corpusname.

%% file: sections/appendix.tex
\section*{Appendix}

\appendix


\titlecontents{section}[18pt]{\vspace{0.05em}}{\contentslabel{1.5em}}{}
{\titlerule*[0.5pc]{.}\contentspage} 


\titlecontents{table}[0pt]{\vspace{0.05em}}{\contentslabel{1em}}{}
{\titlerule*[0.5pc]{.}\contentspage} 

\startcontents[appendix] 
\section*{Table of Contents} 
\printcontents[appendix]{section}{0}{\setcounter{tocdepth}{4}} 

\startlist[appendix]{lot} 
\section*{List of Tables} 
\printlist[appendix]{lot}{}{\setcounter{tocdepth}{1}} 

\startlist[appendix]{lof} 
\section*{List of Figures} 
\printlist[appendix]{lof}{}{\setcounter{tocdepth}{1}} 

\clearpage
\newpage


\section{Related Work} \label{app:related-work}

\subsection{Discourse Parsing Theories} \label{app:discourse-theory}

There are two prominent theories around discourse parsing and structures. The RST theory \cite{MANNTHOMPSON-rst-theory} defines EDUs as clauses (made of subject, object, and predicate). EDUs are then linked to form a discourse tree. The Penn Discourse Treebank developed a parser to divide a text corpus into EDUs and establish relationships between them using the grammar from RST \cite{Penn-discourse-tree}. The Semantic Discourse Representation Theory (SDRT) realizes the need for discourse in AI-based tools dealing with discourse \cite{Asher2005Jun-sdrt}. SDRT defines the theoretical background of discourse relations. The relationship is driven by dynamic logical semantics and a discourse structure.

\subsection{Other Corpora} \label{app:datasets}
\textbf{STAC} \cite{asher-etal-2016-discourse-stac}: The STAC corpus is built on the online game of ``Settlers of Catan''. The game revolves around multiple players with dynamic resources to play and survive on a newly occupied land. Participants interact with each other on a chat system. The interaction includes gameplay interactions and general conversations. Hence, one can replay the entire game by noting the chat interactions. The STAC corpus is built on the recordings of the chat interface and hence includes gameplay-related interactions and general conversations. \citet{asher-etal-2016-discourse-stac} used the SDRT discourse theory to annotate 17 relation types between EDUs.

\noindent \textbf{Molweni} \cite{li-etal-2020-molweni}: The Molweni dataset is based on Ubuntu support chat. This is a multiparty chat environment and is domain-specific. The annotation is based on the SDRT discourse theory and contains 17 relation types between EDUs. 

Table \ref{tab:dataset-comparison} compares the STAC and Molweni datasets with our proposed dataset. \corpusname\ is built by transcribing audio call interactions between a customer and a call center representative. We sourced data from multiple customer call centers catering to domains, including e-commerce, pharmaceutical, stock broker application support, e-marketplace, and education counseling. On the other hand, STAC and Molweni datasets consist of single domains, namely Catan conversations and Ubuntu support. \corpusname\ is built from Hindi-English code-mixed audio conversations with imperfect transcription and diarization quality, imposing a practical outlook on discourse parsing in conversations. 

\subsection{Previous Methods} \label{app:related-model}

\noindent \textbf{Deep Sequential} \cite{deep-sequential}: develops non-structured and structured EDU representations for jointly optimizing link prediction and relation classification. The model sequentially predicts the link and classifies relations for each EDU in a dialog. Glove embeddings are taken for tokens in the EDU and used for downstream models. 

\noindent \textbf{Hierarchical} \cite{liu-chen-2021-improving}: The authors employ a hierarchical text embeddings approach by first encoding the text using a transformer followed by a BiGRU layer to compute EDU representations. Links are predicted by concatenating the representations of an EDU with all the previous EDUs and passing them through a linear layer. A discourse relation is classified by concatenating the representations of two connected EDUs. The authors experiment on STAC and Molweni datasets and highlight a need for domain adaptive models. Since STAC and Molweni are single-domain datasets, they are ineffective in training a model for cross-domain discourse parsing.

\noindent \textbf{Structure-aware} \cite{ijcai2021p543-structure-self-aware} jointly optimizes link and relation prediction. The EDUs are passed through a Hierarchical GRU to obtain context-aware dialog-level embeddings. This is then passed through a GNN containing a structure-aware dot product attention module to compute relation embeddings. As a discourse graph is a DAG, the relation embeddings here are computed for the forward and backward directions. These relation embeddings are then used for link prediction and relation classification.

\noindent \textbf{SSP-BERT+SCIJE} \cite{yu-etal-2022-speaker}: The authors finetune a BERT model to predict if 2 EDUs have the same speaker, which is termed as SSP-BERT. The model then concatenates the embeddings of different speakers and the same speaker using a standard BERT and SSP-BERT model to predict links and classify discourse relation labels jointly.

\noindent \textbf{SDDP} \cite{chi-rudnicky-2022-structured}: This model jointly optimizes link and relation prediction on tree-level distributions. They discard a fraction of the edges to convert the discourse graph from a directed-acyclic graph (DAG) to a minimum spanning tree (MST) to efficiently learn and decode the discourse structure. The discourse tree is learned by minimizing the KL divergence between the predicted and reference tree distributions. The probability distribution of the tree is calculated by computing a tree's score and dividing it by the score of all possible tree structures, i.e., the partition function. The partition function is approximated using the Matrix-Tree theorem \cite{matrix-tree-theorem}.

\section{Corpus Creation} \label{app:corpus-creation}

\subsection{Automatic Speech Recognition (ASR)} \label{app:corpus-asr}

Our ASR system leverages the WavLM model \cite{chen-2022-wavlm-9814838} to generate frame-level embeddings from 8 kHz audio data \cite{verma23-interspeech-asr}. For each 50ms frame, WavLM predicts character probabilities, which are decoded using a beam search algorithm to produce the transcript. To enhance transcription accuracy, we integrate KenLM \cite{heafield-2011-kenlm}, a statistical language model that effectively handles the linguistic diversity of Indian code-mixed speech. The transcription process begins with a reduced character set based on Devanagari, which facilitates phonetic alignment and reduces transcription errors. Subsequently, this text is converted to the native language, where spoken words are mapped to their respective languages. Finally, the text undergoes a romanization process to ensure consistency and maintain the pronunciation of English words, enabling seamless handling of multilingual utterances \cite{verma23-interspeech-asr}.

\subsection{Speaker Diarization} \label{app:corpus-diarization}

We adopt a tailored approach for speaker diarization, addressing both dual-channel and mono-channel audio scenarios. In dual-channel diarization, each speaker’s voice is recorded on a separate channel, and timestamps are assigned to speakers, prioritizing the high-energy speaker in overlapping segments. For mono-channel audio, we employ a clustering-based method using Titanet \cite{koluguri-2021-titanet-neural-model-speaker} to generate embeddings for fixed-length audio windows. By comparing these embeddings with the agent's pre-existing voiceprint, we accurately attribute speech segments to either the agent or the customer.

\section{Annotation details} \label{sec:annotation-details}

\subsection{Annotation Software} \label{app:anno-software}

We used the Inception software \cite{klie-etal-2018-inception} to annotate \corpusname. The software provided the annotators a platform to select the text spans corresponding to an EDU, establish a link between two EDUs, and annotate a relation label for the link. The platform also displayed the description of each annotation label during annotation to remind them of its definition.

\subsection{Annotator Profiles and Payment} \label{app:anno-profile}

The annotators were hired as freelance employees to annotate 20 batches of data for a fixed payment of \$ 1,179.13. Each batch consists of 50 dialogues and consumes 5 hours per annotator. Hence, the annotators were paid \$ 11.79 per hour or \$ 0.60 per dialogue.

\noindent The annotators had previous experience annotating conversation data for various domains, including the domains covered in \corpusname. They are proficient in reading, speaking, and listening to English and Hindi and use both languages in a code-mixed style in everyday communication.

\subsection{Annotation Instructions} \label{app:anno-insturctions}

The annotators were given the following instructions to annotate their batch:

\begin{itemize}[nosep,leftmargin=*]
    \item Dialogue Overview
    \begin{itemize}[nosep,leftmargin=*]
        \item Each dialogue consists of approximately 10 utterances.
        \item An utterance is a sequence of phrases, with each phrase separated by punctuation marks.
        \end{itemize}
    \item Span Identification
        \begin{itemize}[nosep,leftmargin=*]
        \item A span may consist of an entire utterance or one or more phrases within an utterance.
        \item Carefully identify spans where a relation might be possible with another span in the dialogue.
        \end{itemize}
    \item Relation Creation
        \begin{itemize}[nosep,leftmargin=*]
        \item Once relevant spans are identified, create a relational edge between these spans.
        \item Select the appropriate label from the defined relation types to describe the connection.
        \end{itemize}
    \item Edge Constraints
        \begin{itemize}[nosep,leftmargin=*]
        \item No back edges should be created, meaning edges should only flow forward in the dialogue.
        \end{itemize}
    \item Special Instructions on \texttt{Acknowledgment} vs. \texttt{Question-Answer Pair}
        \begin{itemize}[nosep,leftmargin=*]
        \item \texttt{Acknowledgment} is used for statements that function as conversation continuators, indicating understanding.
        \item If an utterance is framed as a question, even if the reply is a simple continuator (e.g., “hmm,” “okay,” “I see”), the relation should be labelled as \texttt{Question-Answer Pair} rather than \texttt{Acknowledgment}.
        \end{itemize}
    \item By following these steps, you will ensure consistent and accurate annotations across the dialogues. Read the entire dialogue first, identify potential relations, mark the spans, and then apply the relevant relation edge labels.
\end{itemize}

\noindent The annotators were also given the list of relation labels, their definitions, and appropriate examples as listed in Table \ref{tab:discourse-labels}.

\subsection{Annotation Process}  \label{app:anno-process} 

A two-party dialogue consists of a list of utterances spoken by two speakers. An utterance is a continuous set of words spoken by a speaker, which may include multiple sentences. The annotators identified elementary discourse units (EDUs) from the utterances for discourse linking and relation labeling. We used clauses as the EDUs based on the definition in Segmented Discourse Relation Theory (SDRT) \cite{Asher2005Jun-sdrt}. 

\noindent Three annotators were recruited to annotate the whole dataset. We divided the dataset into three batches. For each batch, two annotators independently annotated the data, and the third annotator resolved discrepancies. The three batches were rotated in such a way that each annotator annotated two batches and resolved discrepancies on another batch.

\begin{table*}[t]
\centering
\resizebox{0.9\textwidth}{!}{
\begin{tabular}{p{0.2\textwidth}p{0.4\textwidth}p{0.4\textwidth}}
\toprule
\textbf{Discourse Label}        & \textbf{Description} & \textbf{Example} \\
\midrule
\texttt{Acknowledgment} & The tail clause is an agreement or disagreement to the head clause & {\dn jF nAm } confirm {\dn krn\? k\? Ele D\306wyvAd } \newline (\textit{Ok, thank you for confirming your name}) \\ \midrule
\texttt{Question-Answer Pair} & The tail is an answer clause to the question in the head clause & {\dn m\4{\qva} aApkF Eks \3FEwkAr shAytA kr sktA \8{h}\1} $\rightarrow$ {\dn yh m\?rA} return  {\dn k\?} regarding call {\dn h\4} \newline \textit{How can I help you?} $\rightarrow$ \textit{This is a call regarding my return} \\ \midrule
\texttt{Question-Answer Complaint Pair} 
& Similar to the \texttt{Question-Answer Pair}, however, the head clause is a customer complaint question & Sixth {\dn \7{m}J\?} last time {\dn EdKA rhA TA l\?Ekn ab} ninth {\dn EdKA rhA } $\rightarrow$ {\dn hA\1 sr{\rs ,\re} m\?n\?} high priority issue raise {\dn kr EdyA h\4 } \newline (\textit{It was showing me on the sixth, now it is showing ninth} $\rightarrow$ \textit{Yes sir, I have raised a high priority issue}) \\ \midrule
\texttt{Background} & The tail provides supplementary context or information to the subject or object in the head clause. The subject or object in the head clause is the main topic of discussion in the dialogue & {\dn is Evqy m\?{\qva} aApn\?} already issue highlight {\dn EkyA h\4 } $\rightarrow$ 29th October {\dn  kF} date {\dn m\?{\qva} hF} issue highlight {\dn \7{h}aA h\4{\rs ,\re} toh } system {\dn m\?{\qva} } show {\dn ho rhA h\4 } \newline (\textit{You have highlighted an issue regarding this} $\rightarrow$ \textit{The systems shows a issue} \newline \textit{highlighted on 29th October}) \\  \midrule
\texttt{Contrast} & The tail highlights a difference between the subject, predicate, and object interaction in the head clause & {\dn yh} complaint {\dn aAp kr skt\? ho yA \7{m}J\?} online {\dn krnF hogF} $\rightarrow$ {\dn aApko krnF pw\?gF } \newline (\textit{Can you raise the complaint or do I have to do it online?} $\rightarrow$ \textit{You'll have to do it})
 \\ \midrule
\texttt{Correction} & The tail clause is a correction or refinement of the head clause & {\dn aApk\?} headphone {\dn \30CwrAb h\4 } $\rightarrow$ 
{\dn nhF{\qva}{\rs ,\re}} deliver {\dn nhF{\qva} \7{h}e } \newline (\textit{Your headphones are broken} $\rightarrow$ \textit{No, headphones are not delivered}) \\ \midrule


\texttt{Question extension}  \newline (\texttt{Clarification Question}, \texttt{Question elaboration})
& The tail and head are question clauses from the same speaker. The tail enquires more details, seeks clarity, or elaborates on the head clause with option choices. & You are receiving complete wrong item right? $\rightarrow$ Pickup address will be same? \\ \midrule

\texttt{Conditional} \newline (\texttt{Alternation}, \texttt{Conditional})
& The tail provides choices for the actions dictated in the head or sets up a situation that affects the head clause. & ``Either we go now, or we wait for tomorrow'' \newline ``If it rains, we’ll stay inside''  \\ \midrule

\texttt{Continuation} \newline (\texttt{Comment}, \texttt{continuation}, \texttt{elaboration}, \texttt{parallel}, \texttt{result})
& The tail adds a remark, extends or elaborates, clarifies, adds related information, or shows the outcome of a previous action & {\dn \7{s}bh aAyA TA phl\?} message {\dn kF voh EpQ\7{k}\qq{p} k\? Ele Enkl \8{c}kA h\4} agent
$\rightarrow$
{\dn aOr EPr m\?r\? pAs \7{k}C d\?r bAd } second message {\dn aAyA kF EksF} unavoidable event {\dn kF vjh s\?} pickup {\dn nhF{\qva} ho pAy\?gA} \newline
(\textit{I got a message in the morning that the agent has left for receiving the pickup} $\rightarrow$ \textit{Then I got a message saying that the pickup cannot be completed due to an unavoidable event})
 \\
\bottomrule
\end{tabular}
}
\caption[Discourse relation labels and descriptions]{Discourse relation labels and their descriptions. We use a subset of the labels presented in the STAC corpus and add another label, \texttt{Question answer Complaint Pair} to capture a specific case in customer center data. The annotators were given these descriptions and examples during the annotation process. In the first column, we highlight the combined discourse labels for annotating the dataset within parentheses.}
\label{tab:discourse-labels}
\end{table*}

\subsection{Inter Annotator Agreement Metrics}  \label{app:anno-metric}

Table \ref{tab:annotation-metrics} highlights the inter-annotator metrics that we define in Algorithms \ref{alg:exact-match} and \ref{alg:partial-match}. We did not rely on off-the-shelf models and algorithms to segment the text into EDUs because of the nature of \corpusname. It consists of overlapping utterances and imperfect diarizations, which caused segmentation models to split a potentially single EDU into two parts. The annotators were tasked to select the EDU span, build links between EDUs, and classify relation labels. Thus, we calculated the Kappa inter-annotator agreement based on the overlap between the selected spans of each annotator and the links and relation types between EDUs.


\begin{algorithm}[t]
\small
\caption{Span exact match}\label{app:alg-span-exact-match}
\begin{algorithmic}[1]

    
\Require List of spans $A, B$
\Procedure{CountExactMatches}{$A, B$}
    \State $\text{ExactCount} \gets 0$
    \For{$a \in A$}
        \If{$a \in B$}
            \State $\text{ExactCount} \gets \text{ExactCount} + 1$
        \EndIf
    \EndFor
    \State \Return $\text{ExactCount}$
\EndProcedure
\end{algorithmic}
\label{alg:exact-match}
\end{algorithm}

\begin{algorithm}[t]
\small
\caption{Span partial match}\label{app:alg-span-partial-match}
\begin{algorithmic}[1]

\Require List of spans $A, B$ 
\Procedure{CountPartialMatches}{$A, B, \text{threshold}$}
    \State $\text{PartialCount} \gets 0$
    \For{$a \in A$}
        \State $\text{BestMatchScore} \gets \max_{b \in B} \text{Jaccard}(a, b)$
        \If{$\text{BestMatchScore} \geq \text{threshold}$}
            \State $\text{PartialCount} \gets \text{PartialCount} + 1$
        \EndIf
    \EndFor
    \State \Return $\text{PartialCount}$
\EndProcedure
\end{algorithmic}
\label{alg:partial-match}
\end{algorithm}

\begin{figure}[t]
    \centering
    \includegraphics[width=0.6\columnwidth]{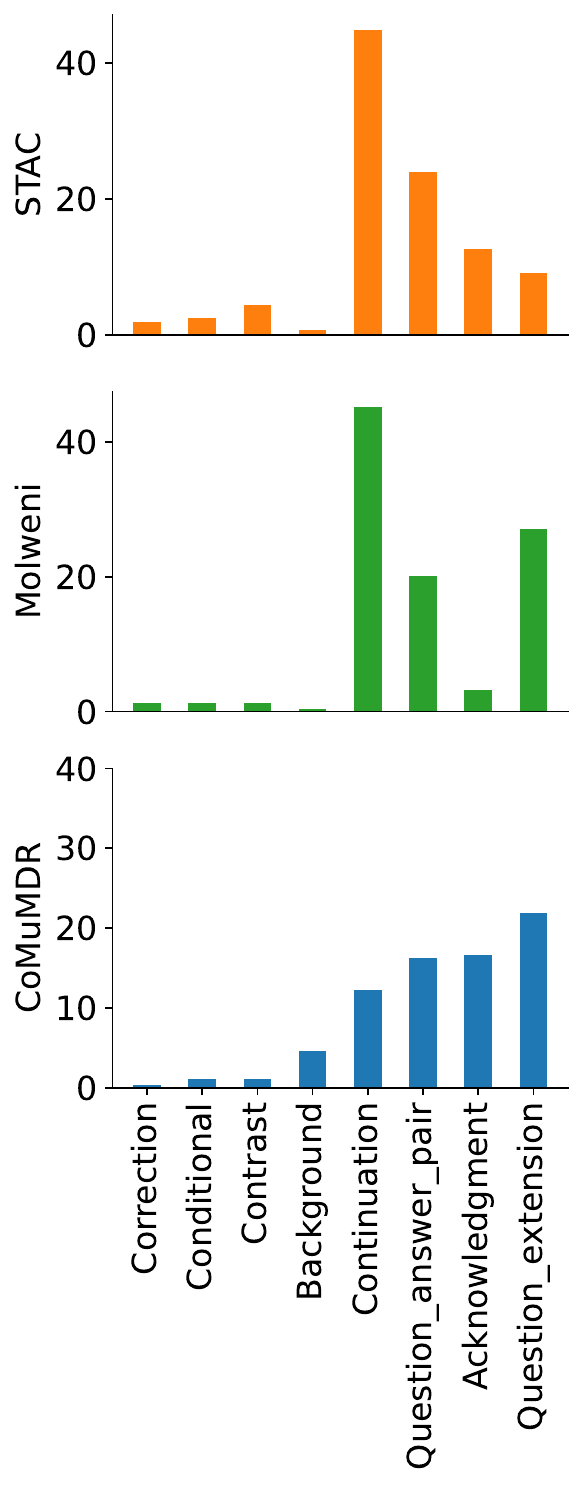}
     \caption[Discourse relation distribution]{Distribution of the discourse relation labels for STAC and Molweni datasets. In this plot, we have combined the labels based on our labeling strategy mentioned in Table \ref{tab:discourse-labels}.}
     \label{fig:stac-molweni-label-distribution}
\end{figure}

\begin{table*}[t]
    \centering
    \begin{tabular}{llcccc}
    \toprule
        \textbf{Model} & \textbf{Optimizer} & \textbf{learning-rate} & \textbf{lr-decay} & \textbf{epochs} & \textbf{batch size} \\
        \midrule
        Deep Sequential & AdamW & 1e-1 & 0.98 & 50 & 4 \\
        Hierarchical & AdamW & 2e-4 & 1.00 & 20 & 1 \\
        Structure-aware & SGD & 1e-1 & 0.98 & 10 & 1 \\
        SSP-BERT SCIJE & Adam & 1e-3 & 0.75 & 100 & 4 \\
        SDDP & AdamW & 2e-5 & 1e-8 & 3 & 4 \\
        \bottomrule
    \end{tabular}
    \caption[Hyperparameter settings]{Hyperparameter settings used to experiment all the discourse parsing models on STAC, Molweni, and \corpusname\ datasets.}
    \label{tab:baseline-hyperparams}
\end{table*}

\section{Model Training Details} \label{app:hyper-params}

We used the same hyperparameter settings as mentioned in the model papers. All the experiments were carried out on an Nvidia 3090 GPU. We mentioned the relevant hyperparameters in Table \ref{tab:baseline-hyperparams}. 

\begin{figure}[t]
    \centering
    \includegraphics[width=\columnwidth]{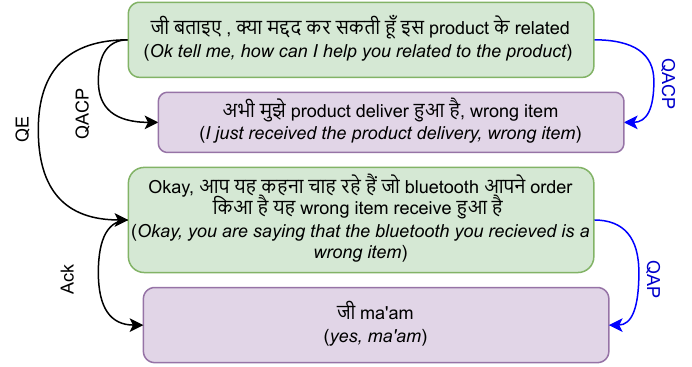}
    \caption[\corpusname\ sample conversation]{A sample conversation taken from \corpusname. Utterances from the customer are marked in purple, and those of the customer center representative are green. The gold and predicted relations are marked on the left and right sides.}
    \label{fig:error-sample-1}
\end{figure}

\section{Evaluation Metric} \label{app:eval-metric}

We compute link prediction as a binary classification task between two EDUs. If a link is present in the gold annotations and prediction, it is a \texttt{True positive} link. Similarly, if a link is predicted between two EDUs but is not in the gold annotation, it is a \texttt{False positive} link. Using these definitions, we construct the confusion matrix and calculate the F1-score for link prediction.

\noindent A relation $r_{ji}$ between two EDUs $(u_j, u_i)$ is classified only if the model predicts a link between $(u_j, u_i)$. Hence, we first find all the intersecting links between the gold annotated data and predicted links, i.e., $\forall j, i$ if there is a link $u_j$ and $u_i$ in gold and predicted data then capture the gold and predicted relations $(r_{ji}, r^\prime_{ji})$. We calculate the link + relation F1-score by using the pairs of gold and predicted relations.


\section{GPT-4 Template} \label{app:gpt-4-template}
We experimented with using GPT-4 for discourse parsing on STAC, Molweni, and \corpusname. We used the prompt template mentioned in Figure \ref{fig:gpt4-prompt}.

\begin{figure}[t]
    \centering
    \begin{tabular}{p{0.9\columnwidth}}
         \toprule
         \texttt{You are given a dialogue conversation between an agent and a customer. You have to do the link and relation prediction using SDRT format. You will be given the relations and you have to stricty use those relations only to do the prediction. You will be given the nodes as well in the form of extracted text spans. During link prediction, you have to identify between which nodes there exists a link and what would be the relation.\newline \newline
         you have to return the answer in the SDRT format like json. Do not return any extra text or explanation. \newline \newline
         Dialogue: \newline
         \{dia\} \newline \newline
         Spans: \newline
         \{spans\} \newline \newline
         relations: \newline
         \{rels\} \newline \newline
         Following is just an example of annotation: \newline
         \{examples\}
         \newline \newline
         Note: For all the instances where a sentence spoken by the same person is broken down into multiple lines, then use \texttt{dia-continuation} relation.} \\
         \bottomrule
    \end{tabular}
    \caption[GPT-4 prompt template]{Prompt template used for evaluating GPT-4 as a discourse parser.}
    \label{fig:gpt4-prompt}
\end{figure}


\begin{table*}[h]
    \centering
    \begin{tabular}{lrrr}
    \toprule
    \textbf{Relation} & \textbf{STAC} & \textbf{Molweni} & \corpusname \\
    \midrule
    \texttt{Continuation}                    & 1.17 $\mp$ 1.53 & 1.65 $\mp$ 1.14 & 1.06 $\mp$ 0.42 \\
    \texttt{Question answer}                 & 1.78 $\mp$ 1.20 & 1.56 $\mp$ 1.09 & 0.99 $\mp$ 0.27 \\
    \texttt{Acknowledgment}                 & 1.67 $\mp$ 1.31 & 1.41 $\mp$ 0.81 & 0.95 $\mp$ 0.32 \\
    \texttt{Background}                      & 0.72 $\mp$ 1.14 & 1.35 $\mp$ 0.66 & 1.07 $\mp$ 0.39 \\
    \texttt{Correction}                      & 1.67 $\mp$ 1.84 & 1.41 $\mp$ 0.77 & 1.00 $\mp$ 0.00 \\
    \texttt{Question Extension}              & 1.86 $\mp$ 1.86 & 1.62 $\mp$ 1.12 & 1.05 $\mp$ 0.48 \\
    \texttt{Conditional}                     & 0.67 $\mp$ 1.35 & 1.78 $\mp$ 1.33 & 1.03 $\mp$ 0.33 \\
    \texttt{Contrast}                        & 0.97 $\mp$ 1.15 & 1.60 $\mp$ 1.05 & 0.98 $\mp$ 0.19 \\
    \texttt{Question answer complaint}       & - & - & 1.21 $\mp$ 0.54 \\
    \texttt{dia-continuation}                & - & - & 0.97 $\mp$ 0.26 \\
    \bottomrule
    \end{tabular}
    \caption[EDU distance distribution]{Mean and standard deviation of distribution of distance between linked EDUs for all corpus}
    \label{tab:relation_distance_distribution}
\end{table*}

\section{Distance between linked EDUs} \label{app:distance-edus}

\begin{figure*}[tbp]
    \centering
    \includegraphics[scale=0.70]{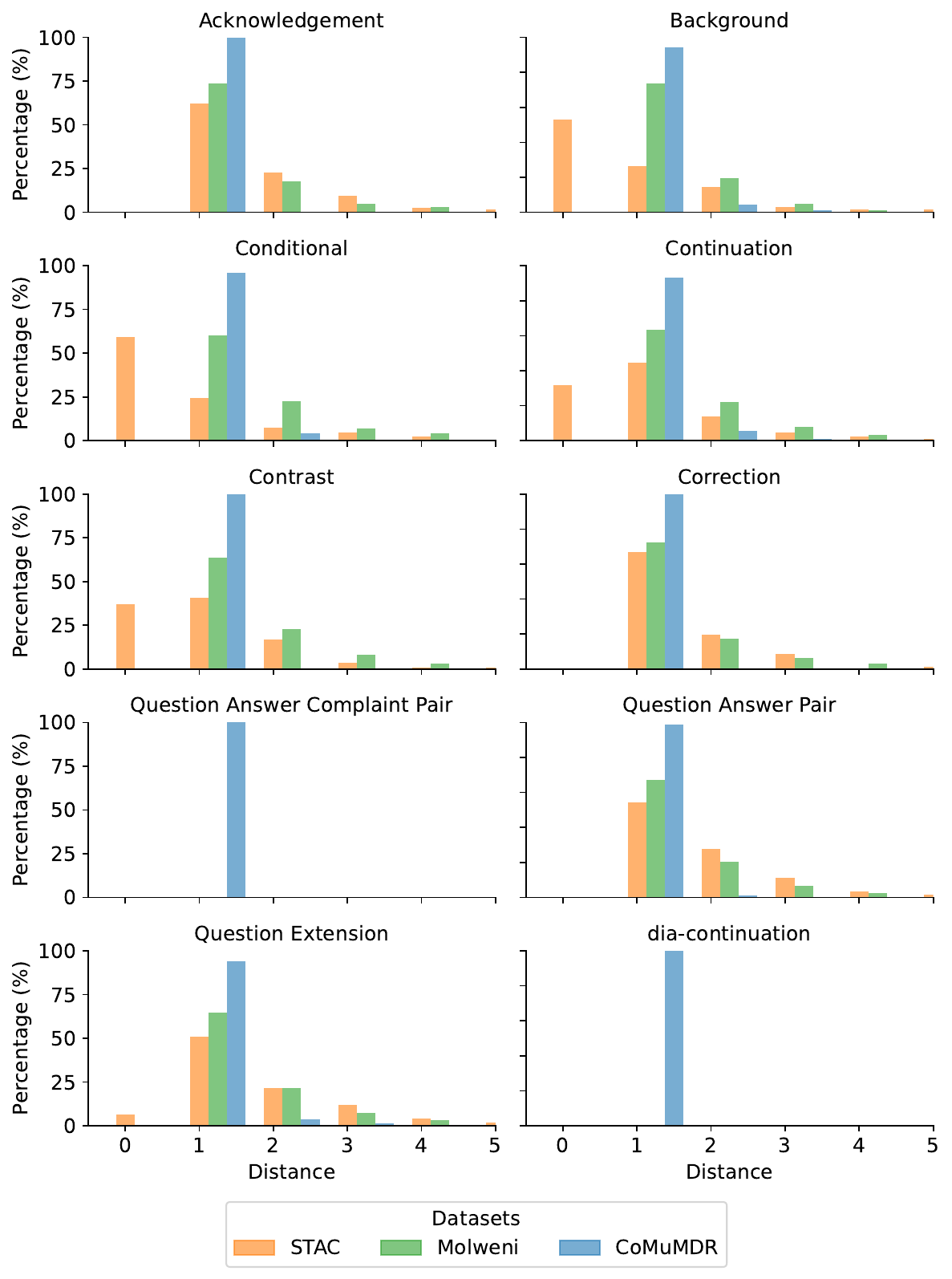}
    \caption{Distance between linked EDUs for different corpora}
    \label{fig:edu-distance}
\end{figure*}

Fig. \ref{fig:edu-distance} (and Table \ref{tab:relation_distance_distribution}) shows the distribution of relative distance between linked EDU pairs for each relationship type. The distance between EDU pairs is defined as the difference between the utterance turn of the head and tail of a link. For example, a link to the same utterance has 0 distance while a link to the next utterance has 1 distance. We merged the statistics of the merged relations (mentioned in Table \ref{tab:discourse-labels} for STAC and Molweni. We observe a significant distribution overlap between STAC and Molweni datasets for \texttt{Correction}, \texttt{Question Extension}, \texttt{Acknowledgment} and \texttt{Question\_answer\_pair}, suggesting their relative similarity. However, for \texttt{Conditional}, \texttt{Continuation} and \texttt{Contrast} there is a difference in the distributions. We also plot the same for \corpusname. We notice a significant difference in the distributions; notably that most of the relations have a distance of 1. We also look at the mean and standard-deviation of the relation distances in Table \ref{tab:relation_distance_distribution}. The median distance between linked EDUs for all relations is 1 in all the datasets. 



\begin{table*}[ht]
    \centering
    \resizebox{\textwidth}{!}{
    \begin{tabular}{l|cccccccccc}
        \toprule
        ~ & \texttt{\textbf{Ack}} & \textbf{\texttt{dc}} & \textbf{\texttt{QAP}}  & \texttt{\textbf{QACP}} & \textbf{\texttt{QE}} & \textbf{\texttt{Correction}} & \textbf{\texttt{Continuation}} & \textbf{\texttt{Conditional}} & \textbf{\texttt{Background}} & \textbf{\texttt{Contrast}} \\
        \midrule
        \textbf{\texttt{Ack}} & 59 & 28 & 3 & 0 & 0 & 0 & 3 & 0 & 3 & 1 \\ 
        \textbf{\texttt{dc}} & 20 & 70 & 10 & 1 & 2 & 4 & 27 & 0 & 7 & 1 \\ 
        \textbf{\texttt{QAP}}  & 28 & 17 & 42 & 0 & 1 & 6 & 15 & 1 & 0 & 0 \\ 
        \textbf{\texttt{QACP}} & 1 & 0 & 1 & 0 & 0 & 0 & 0 & 0 & 0 & 0 \\ 
        \textbf{\texttt{QE}} & 11 & 20 & 15 & 1 & 14 & 3 & 28 & 3 & 3 & 2 \\ 
        \textbf{\texttt{Correction}} & 0 & 0 & 0 & 0 & 0 & 1 & 0 & 0 & 0 & 0 \\ 
        \textbf{\texttt{Continuation}} & 9 & 20 & 1 & 0 & 4 & 2 & 26 & 1 & 4 & 3 \\ 
        \textbf{\texttt{Conditional}} & 0 & 2 & 0 & 0 & 0 & 0 & 1 & 1 & 1 & 0 \\ 
        \textbf{\texttt{Background}} & 2 & 9 & 0 & 0 & 0 & 0 & 7 & 0 & 3 & 0 \\ 
        \textbf{\texttt{Contrast}} & 0 & 1 & 1 & 0 & 0 & 3 & 2 & 0 & 0 & 0 \\ 
        \bottomrule
    \end{tabular}
    }
    \caption[GPT-4o confusion matrix]{Confusion matrix of discourse link+relation classification done by GPT-4o. We have turned some relations into their relevant acronyms for viewing: \texttt{Ack}-\texttt{Acknowledgment}, \texttt{QACP}-\texttt{Question Answer Complaint Pair}, \texttt{QAP}-\texttt{question answer pair}, \texttt{QE}-\texttt{question extension}, and \texttt{dc}-\texttt{diarization continuation}.}
    \label{tab:gpt-confusion-matrix}
\end{table*}